\documentclass[sigconf]{acmart}

\AtBeginDocument{%
  \providecommand\BibTeX{{%
    \normalfont B\kern-0.5em{\scshape i\kern-0.25em b}\kern-0.8em\TeX}}}

\author{Lilas Alrahis, Johann Knechtel, and Ozgur Sinanoglu}
\email{{lma387, jk176, os22}@nyu.edu}
\affiliation{%
  \institution{New York University Abu Dhabi}
  \country{Abu Dhabi, United Arab Emirates (UAE)} 
}

\renewcommand\footnotetextcopyrightpermission[1]{}

\usepackage{soul}
\newcommand{\drop}[1]{\textcolor{red}{#1}}
\renewcommand{\drop}[1]{}
\definecolor{cadmiumgreen}{rgb}{0.0, 0.42, 0.24}
\usepackage{multirow}
\usepackage{enumitem}

\usepackage{enumitem}
\usepackage{graphicx}
\usepackage{booktabs} 
\usepackage{pifont}
\usepackage{mathtools}
\def\smallerspacecaption{\vspace{-2mm}}

\usepackage[ruled, vlined, norelsize]{algorithm2e}
\SetKwInput{KwInput}{Input}
\SetKwInput{KwOutput}{Output}
\SetKwInput{KwInit}{Initialization}
\SetKwInput{Kwprocedure}{Procedure}
\usepackage{lipsum}
\usepackage{dblfloatfix}

\usepackage{bm}

\usepackage{setspace}
\usepackage{scalerel}[2016/12/29]

\definecolor{green}{HTML}{2a7e6b}

\newcommand{\ssub}[2]{{#1}_{\scaleobj{0.8}{#2}}}

\usepackage{blindtext,graphicx}

\usepackage{bm}
\usepackage{circledsteps}
\pgfkeys{/csteps/inner color=white}
\pgfkeys{/csteps/outer color=none}
\pgfkeys{/csteps/fill color=black}

\usepackage{algorithmic}
\usepackage{textcomp}
\AtBeginDocument{%
  \providecommand\BibTeX{{%
    \normalfont B\kern-0.5em{\scshape i\kern-0.25em b}\kern-0.8em\TeX}}}

\usepackage{multirow}
\usepackage{graphicx}

\usepackage{amsmath,amsfonts,bm}

\def\mA{{\bm{A}}}

\def\mz{{\bm{z}}}

\def\mX{{\bm{X}}}

\def\mZ{{\bm{Z}}}

\DeclareMathAlphabet{\mathsfit}{\encodingdefault}{\sfdefault}{m}{sl}
\SetMathAlphabet{\mathsfit}{bold}{\encodingdefault}{\sfdefault}{bx}{n}

\def\gG{{\mathcal{G}}}

\def\gN{{\mathcal{N}}}

\newcommand{\R}{\mathbb{R}}

\begin{document}

\title{Graph Neural Networks: A Powerful and Versatile Tool for Advancing Design, Reliability, and Security of ICs}

\begin{abstract}

Graph neural networks (GNNs) have pushed the state-of-the-art (SOTA) for performance in learning and predicting on
large-scale data present in social networks, biology, etc.
Since integrated circuits (ICs) can naturally be represented as graphs, there has been a tremendous
surge in employing GNNs for machine learning (ML)-based methods for various aspects of IC design. Given this
trajectory, there is a timely need to review and discuss some powerful and versatile GNN approaches for
advancing IC design.

In this paper, we propose a generic pipeline for tailoring GNN models toward solving challenging problems for IC
design. We outline promising options for each pipeline element, and we discuss selected and promising works, like
leveraging GNNs to break SOTA logic obfuscation.
Our comprehensive overview of GNNs frameworks covers (i) electronic design automation (EDA) and IC design in general,
(ii) design of reliable ICs, and (iii) design as well as analysis of secure ICs.
We provide our overview and related resources also in the {GNN4IC} hub at \url{https://github.com/DfX-NYUAD/GNN4IC}.
Finally, we discuss
interesting open problems for future research.

\end{abstract}

\maketitle
\renewcommand{\headrulewidth}{0.0pt}
\thispagestyle{fancy}
\lhead{}
\rhead{}
\chead{\copyright~2023 ACM.
This is the author's version of the work. It is posted here for personal use.
Not for redistribution.	The definitive Version of Record is published in ASP-DAC 2023.}
\cfoot{}
\section{Introduction}
\label{sec:intro}
The increasing complexity of IC design and manufacturing, coupled with the advancements in ML, have led researchers to
employ various ML methodologies to advance different aspects of IC design. Accordingly, ML has beaten SOTA approaches
in solving various EDA tasks, including hardware security and reliability. Among the different ML methodologies
explored, GNNs have demonstrated great potential in transforming IC design workflows.

Circuits, whether in register-transfer level (RTL), gate-level or transistor-level netlist, or layout format, can be naturally
represented as graphs. Thus, most EDA tasks can be modelled as graph problems. Given a rapid development of
GNNs, the past three years have witnessed an exponential surge of research interest in using GNNs to address various
circuit-related tasks.
With many options available for circuit-to-graph conversion, extraction, GNN architectures -- all for varying
tasks --
it is vital to summarize the achievements in the field and point out paths for future research.

There are few if any reviews on GNNs covering multiple domains of IC design.
References~\cite{MLCAD,gcn,huang21,ma2020understanding} present surveys on ML for advancing EDA in general,
whereas
References~\cite{9531070,10.1145/3543853,ren22,lu22} represent some up-to-date
surveys on GNNs used for EDA. Wu~\textit{et~al}~\cite{10.1145/3489517.3530408} survey the usage of GNNs for high-level synthesis (HLS) performance prediction. For hardware security, Liu~\textit{et~al.}~\cite{boonsbanes} and Sisejkovic~\textit{et~al.}~\cite{ml_locking_survey} review the use of ML in general,
whereas Alrahis~\textit{et~al.}~\cite{alrahis2022embracing} present a thorough review for using GNNs in particular.
In this paper, we cover a wider range of important topics, presenting selected papers on
GNNs for IC design, reliability, and security. The contributions of this paper are as follows.
\vspace{-0.1mm}
\begin{enumerate}[leftmargin=*]
\item We present a \textbf{taxonomy and a generic pipeline of GNNs} for advancing IC design, mainly (i) GNNs for EDA and IC design in
general, (ii) GNNs for reliable IC design, and (iii) GNNs for design and analysis of secure ICs.
\item We provide a \textbf{comprehensive survey}.
We review selected SOTA and provide detailed descriptions on the GNN tasks, architectures, graph types, circuit design level, features, etc.
\item We collect resources on GNNs for IC design, including the different applications, GNN models, benchmark
datasets, and open-source codes, and summarize them in our \textbf{\textit{GNN4IC} hub} [\url{https://github.com/DfX-NYUAD/GNN4IC}].
\end{enumerate}

\section{Background}
\label{sec:background}
\subsection{Electronic Design Automation}
\label{sec:background_EDA}
Sophisticated EDA tools are essential
for IC designers to utilize advances in microelectronics
while managing the ever-increasing complexities at the
same time.
In recent years, ML has become a key driver for next-generation EDA tools, as we also show in this
paper.
Still, the overall EDA pipeline remains largely the same.

\textbf{System specification and architectural design.} The objectives and high-level requirements for
functionality, performance, and physical implementation are formulated.
Modeling languages like SystemC can be used for a formalized description.

\textbf{Behavioral and logic design.} The specification and architecture are transformed into a behavioral RTL model which
describes inputs, outputs, timing behavior, etc. for the whole system. 
Third-party components can be integrated at this stage as needed. Traditionally,
this step is done manually, but nowadays it is also well supported by EDA tools offering HLS.

\textbf{Logic synthesis.} The behavioral model is transformed into a low-level circuit description, the gate-level
netlist (GLN). This step requires the technology library for mapping from the generic model to the specific IC design that
is to be implemented.

\textbf{Physical design.} The GLN is transformed into the actual physical layout of gates, macros, wires,
memories, etc. Considering the high complexity of this stage, it is typically divided into the following tasks: partitioning and/or floorplanning, power and ground delivery, placement, clock delivery, routing,
and timing closure.

\textbf{Verification and signoff.} The physical layout must be verified against various design and
manufacturing rules, to ensure correct functionality and electrical behaviour. Once all rules are met, the design can
be signed off and taped out.

\subsection{Hardware Reliability}
\label{sec:background_rel}
Technology scaling makes it challenging to ensure the reliability of ICs over the projected device lifetime. In the
following, we provide an overview of important reliability concerns.

\textbf{Process variation} occurs due to imperfections in the IC manufacturing process. It is a
time-independent source of variation that differs for each IC and even within the chip itself. Random
dopant fluctuation, metal gate granularity, and line-edge roughness are among the typical sources of variation that
are considered for SOTA FinFET devices~\cite{bib:zhang2018extraction}. An accurate estimation of process variation is
prerequisite to ensure high yield at high performance.

\textbf{Transistor aging} is a time-dependent source of variation that depends on 
many parameters such as temperature, workload, and projected lifetime. Physical effects that lead to transistor aging
include hot-carrier injection (HCI), and importantly, bias temperature
instability~(BTI)~\cite{bib:amrouch2016reliability}. BTI is one of the dominant contributors to aging-induced
degradations. Over an IC's lifetime, electrical charges get trapped in the gate oxide of transistors,
resulting in increased threshold voltages. In addition, interface traps can be generated at the $Si$-$SiO2$
layer resulting in more undesired charges and, hence, further increase in the threshold voltage. Consequently,
transistor switching times and circuit path delays increase.

\subsection{Hardware Security}
\label{sec:background_security}

The IC manufacturing process has
become globalized, involving numerous entities across the globe.
Such an outsourced supply chain leads to a plethora
of security concerns, like implantation of malicious hardware Trojans (HTs), piracy of design intellectual
property (IP), and reverse-engineering (RE)~\cite{DfTr_techniques_IEEE_proceedings,rostami2014primer}, any of which
can cause financial loss to IP owners and/or put end users at risk.

Next, we review selected threats and countermeasures relevant for this paper at hand.
References~\cite{knechtel2019protect,knechtel2020towards,knechtel20survey}
provide a wider coverage, also touching upon how to advance EDA for secure IC design.

\subsubsection{Hardware Security Threats}

\textbf{HTs} are malicious modifications to ICs that attackers implant to achieve a malignant
outcome~\cite{rostami2014primer}, such as leaking sensitive assets, e.g., crypto keys, or disrupt design
functionality. HTs are typically stealthy (i.e., rarely activated) and can be integrated at different stages of the IC
design flow, e.g., through adversarial third-party IP cores, ``hacked'' design tools, or by
malign employees during chip fabrication~\cite{knechtel20survey}.

\textbf{IP piracy} refers to the theft of the design IP by an adversary (e.g., foundry or end-user) to develop
competing devices without incurring the research and development costs~\cite{knechtel20survey}.

\textbf{RE} is a process that aims to obtain the IC/IP design, technology, or functionality by analyzing the
chip layer by layer~\cite{mcloughlin2011reverse}. RE can be leveraged either for various
attacks or as defense approach, e.g., to detect IP infringement, verify IP implementation, and detect HTs.

\subsubsection{Design-for-Trust Methods}

\textbf{Layout camouflaging} alters the appearance of a chip to conceal the design IP, i.e., it obfuscates the design
information either at the transistor-level~\cite{patnaik2018advancing}, gate-level~\cite{rajendran2013security}, or
interconnect-level~\cite{patnaik2020obfuscating}, protecting it from RE attacks.

\textbf{Logic locking} obfuscates the structure and functionality of a design by integrating key-controlled logic elements,
referred to as \textit{key-gates}. These key-gates bind the correct functionality of the design to a secret key that
is only known to the legitimate IP owner.
The owner loads the secret key into an on-chip, tamper-proof memory
after fabrication and testing.
References~\cite{Subramanyan_host_2015,unsail,scansat_aspdac,redundancy,scansat_journal,alrahis2019functional,snapshot}
provide further details on logic locking.

\begin{figure*}[!t]
\centering
\includegraphics[width=0.99\textwidth]{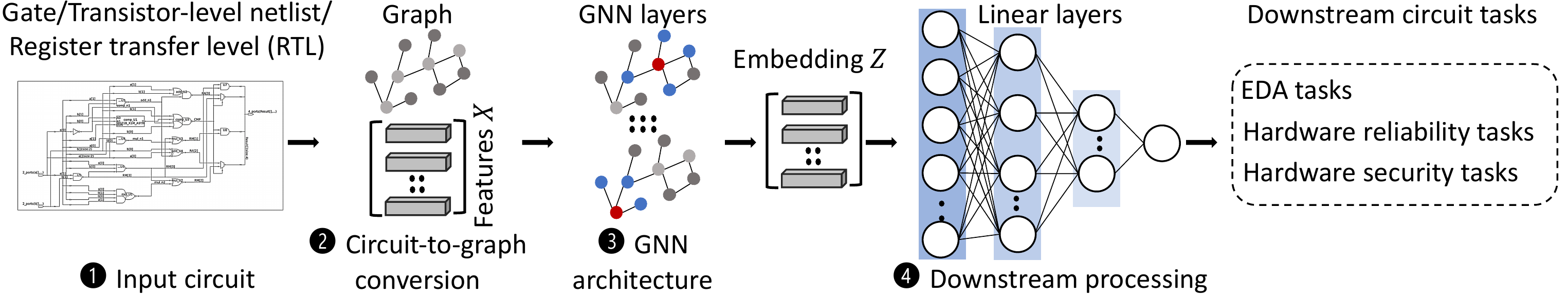}
\smallerspacecaption
\caption{Generic end-to-end pipeline for using GNNs to tackle circuit-related tasks.}
\label{fig:gnn_model}
\end{figure*}

\subsection{Graph Neural Networks}
\label{sec:background_gnn}

\begin{table}[!t]{\footnotesize
\renewcommand\arraystretch{0.9}
\centering

\caption{Symbols and notations used in this paper.\label{tab:symbol}}
}
\smallerspacecaption
\resizebox{\columnwidth}{!}{%
\begin{tabular}{llll}
\hline
\textbf{Notation} & \textbf{Definition} & \textbf{Notation} & \textbf{Definition} \\
\hline
$\gG, \ssub{y}{\gG}$ & Graph, class & $\mX$ & Node features matrix \\\hline
$v, \ssub{y}{v}$ & Node, class &$\mz_{\gG}$ & Graph-level embedding\\\hline
 $\mA$ & Adjacency matrix & $h$ & hop-size \\\hline
 $V$ & Set of nodes in $\gG$ & $S$ & Set of target nodes \\\hline
 $E$ & Set of edges in $\gG$ & $d(u,v)$ & Shortest distance b/w $u$, $v$ \\\hline
 $n$ & Number of nodes in $\gG$ & $\gG_{(S,h)}$ & Subgraph from $\gG$ around $S$ \\\hline
 $k$ & Length of feature vector & $\hat{y}$ & Predicted outcome \\\hline
 $\gN(v)$ & Neighbors of nodes $v$ & $g$ & Downstream classifier \\\hline
 $L$ & Number of GNN layers & $\theta^{(l)}$& GNN trainable parameters \\\hline
 $\mZ$ & Node embeddings matrix & $\sigma(.)$ & Activation function \\\hline
\end{tabular}}
\end{table}

Next, we provide a background on GNNs, define important graph-related concepts, and depict the notations used in this paper (Table~\ref{tab:symbol}).
We begin by defining a graph as follows.

\textit{Definition~1.} $\gG=(V,E)$ denotes a \textbf{graph} with set $V$ of nodes and set $E \subseteq V\times V$ of edges. $\mX \in \R^{n \times k}$ is a matrix of node features, where $n$ is the number of nodes in $\gG$. $\mA \in \{0,1\}^{n\times n}$ represents the adjacency matrix of $\gG$ with $\mA_{i,j}=1$ iff $(i,j)\in E$. 

\textit{Definition~2.} A \textbf{directed graph} is a graph in which the edges have a direction from one node to another. An
\textbf{undirected graph} is a special case of directed graphs where there is a pair of edges with opposite directions for every two connected nodes. $\mA$ of an undirected graph is symmetric. A \textbf{hypergraph} is a graph in which an edge can join any number of nodes.

\textit{Definition~3.} \textbf{Heterogeneous graphs} have 
nodes and/or edges of different types, while
\textbf{homogeneous graphs} have nodes and edges of same types.

\textit{Definition~4.} \textbf{GNNs} learn on the structure and node attributes of $\gG$ to generate a representation (\textit{i.e.,} \textit{embedding}) $\mZ$ that facilitates graph-based ML tasks. More specifically, a GNN takes as input a graph $\gG$ and generates an embedding $\mz_{v}$ for each node $v \in V$. 
The GNN updates the node embeddings through multiple iterations of \textit{neighborhood aggregation}, as follows.

\begin{equation}
\mZ^{(l)} = \texttt{Aggregate}\left(A, \mZ^{(l-1)}; \theta^{(l-1)} \right)
\end{equation}
where $\mZ^{(l)}$ is the node embeddings matrix at the $l$-th iteration and $\theta^{(l-1)}$ is a trainable weight matrix. 
$\mZ^{(0)}$ represents the initial node features $\mX$. 
The $\texttt{Aggregate}$ function is typically an order invariant function, such as $\texttt{sum}$, $\texttt{average}$, or $\texttt{max}$.

After $L$ iterations of neighborhood aggregation, the generated node embeddings $\mZ$ can be used directly for
different downstream tasks. That is, a GNN models a function $f_{\theta}$ that generates $\mz_{v} = f_{\theta}(\gG,v)$
for $v \in \gG$, and the embedding is then passed to a downstream model $g$ for classification, regression, or clustering. The predicted outcome for $v \in \gG$ is denoted as $\hat{y}_{v}$, where $\ssub{\hat{y}}{v}=g(\mz_{v})$.

In graph-level tasks, $\texttt{pooling}$ and $\texttt{readout}$ functions generate a graph-level embedding, $\mz_{\gG}$, which can be used for graph classification, regression or matching. Overall, a GNN models a function $f_{\theta}$ that generates $\ssub{z}{\gG} = f_{\theta}(\gG)$. The embedding is then passed to a downstream model $g$ for classification, regression, or matching~\cite{kipf2016semi}. 
The predicted outcome for $\gG$ is denoted as $\hat{y}_{\gG}$, where $\ssub{\hat{y}}{\gG}=g(\mz_{\gG})$. The node embeddings can further be used to solve edge-level tasks, such link prediction and edge classification, which require the model to predict whether there is an edge between two target nodes (identified by set $S$) or to classify edge types.

In GNN-based link prediction, an enclosing subgraph around each target link is extracted~\cite{SEAL}. Given $(S,\gG)$, the $\gG_{(S,h)}$ subgraph induced from $\gG$ by $\cup_{v\in S} \{u ~|~ d(u,v) \leq h\}$, where $d(u,v)$ is the shortest path distance between nodes $u$ and $v$. The subgraphs hold information about the network surrounding the links. Therefore, by performing graph classification, the labels of the target links also become the labels of their corresponding subgraphs.

\section{Proposed Taxonomy}
\label{sec:taxonomy}

Figure~\ref{fig:gnn_model} demonstrates a generic pipeline for using GNNs to solve circuit-related tasks. First, some
circuit, digital or analog, given in RTL, GLN, or transistor-level format, is represented as a graph. Different tasks
require different graph representations, such as directed, undirected, hypergraphs, heterogeneous, or homogeneous
graphs. Various node and/or edge attributes can be extracted from the circuit, depending on the target task. Next, the 
graphs are passed to the GNN model, which solves some ML tasks, such as node classification or regression, graph matching or clustering, etc.

Researchers have utilized various GNN architectures like the graph convolutional network (GCN)~\cite{kipf2016semi}, message
passing neural network (MPNN)~\cite{mpnn}, GraphSAGE~\cite{hamilton2017inductive}, graph isomorphism network
(GIN)~\cite{xu2018powerful}, deep graph convolutional neural network (DGCNN)~\cite{zhang2018end}, graph attention
network (GAT)~\cite{velivckovic2017graph}, principal neighborhood aggregation (PNA)~\cite{corso2020principal}, etc. Researchers may also devise some custom GNN architecture as needed, specific to
the circuits and tasks at hand.
In general, the GNN model can be integrated with other ML algorithms, such as long-short-term memory (LSTM) models and
reinforcement learning (RL).

In the following, we give a detailed survey for the circuit-related domains covered in our taxonomy: EDA, reliability, and security.

\subsection{GNNs for EDA}

GNNs have been proposed for various tasks throughout the EDA pipeline. We review selected works below and summarize in
Table~\ref{tab:GNN_EDA}.

\begin{table*}[tb]
\caption{Summary of selected GNN frameworks for EDA. $-$ means unspecified, and NA means not applicable.}
\smallerspacecaption
\label{tab:GNN_EDA}
\resizebox{\textwidth}{!}{%
\setlength\tabcolsep{2.2pt} 
\begin{tabular}{llllllllll}
\hline
\textbf{Platform} & \textbf{GNN Task} & \textbf{GNN} & \textbf{\#Layers} & \textbf{Loss Function} & \textbf{Graph Type} & \textbf{Circuit Type} & \textbf{Features} & \textbf{Pooling} & \textbf{Readout} \\ \hline
\textbf{\cite{lopera22_ICCAD}} & Edge regression & \begin{tabular}[c]{@{}l@{}}GCN~\cite{kipf2016semi}\\MPNN~\cite{mpnn}\end{tabular} & 3 & MSE & DAG & RTL & \begin{tabular}[c]{@{}l@{}}RTL node type, \#Inputs/Outputs\\ Bitwidth, Depth level\\Edge inverter, Input slew (edge) \\ Output slew delay (edge) \end{tabular}& NA & NA \\\hline

\textbf{D-SAGE~\cite{ustun2020accurate}} & \begin{tabular}[c]{@{}l@{}}Node/Edge\\ classification\end{tabular} & \begin{tabular}[c]{@{}l@{}}Custom \\GraphSAGE\cite{ustun2020accurate}\end{tabular}  & 2 & Cross-entropy& DFG & HLS &\begin{tabular}[c]{@{}l@{}}Operation type\\ Bitwidth \end{tabular}&  NA & NA  \\\hline

\textbf{\cite{zhou22}}& Imitation learning & HGNN~\cite{10.1145/3054912} & 3-6 & \begin{tabular}[c]{@{}l@{}}Weighted\\cross-entropy\end{tabular}& \begin{tabular}[c]{@{}l@{}}Heterogeneous\\directed\end{tabular} & GLN & \begin{tabular}[c]{@{}l@{}}
Footprint, Sizable, Output pin (pin)\\
Fanin, Area, Slack (pin), Cap. (pin)\\
Gain (pin), Slew (pin)\\Delay (edge), RC delay (edge), $\lambda$ (edge)\end{tabular} & NA & NA \\\hline
 
\textbf{CongestionNet~\cite{kirby2019congestionnet}} & Node regression & GAT~\cite{velivckovic2017graph}& 16 & MSE& Undirected & GLN &Cell type, size, Pin count & NA&NA\\ \hline

\textbf{\cite{ghose21}} & Node regression & GraphSAGE~\cite{hamilton2017inductive}& 2 & Squared error& Undirected & GLN & Pin number, Cell size& - & - \\\hline

\textbf{TP-GNN\cite{tpgnn}}& Node clustering & GraphSAGE~\cite{hamilton2017inductive} & 2 & Unsupervised & Undirected clique & GLN & \begin{tabular}[c]{@{}l@{}}Hierarchy, Sums slack, delay, slew\\Shortest path to clock\\\#1- and 2-hop neighbors \end{tabular} & NA & NA\\\hline
\textbf{\cite{lu21}} & Node clustering &GraphSAGE~\cite{hamilton2017inductive} & 2& Unsupervised &Undirected clique & GLN & Hierarchy and memory features & NA & NA \\\hline
\textbf{Edge-GNN~\cite{mirhoseini2021graph}} & GNN-RL & GCN~\cite{kipf2016semi} & 1 &  MSE & Hypergraph &GLN &\begin{tabular}[c]{@{}l@{}}Type, Width, Height\\x and y coordinates\end{tabular} &NA& \begin{tabular}[c]{@{}l@{}}Reduce\\ mean\end{tabular}\\ \hline

\textbf{Net$^{2}$~\cite{net2}} & Node regression & Custom GAT~\cite{net2} & 3 &-& Directed & GLN &  \begin{tabular}[c]{@{}l@{}} Driver's area, Fanin, Fanout \\Clusrer ID (edges)\\\end{tabular} & NA & NA \\\hline

\textbf{GRANNITE~\cite{GRANNITE}} & Node regression & \begin{tabular}[c]{@{}l@{}}Sequential \\GCN~\cite{kipf2016semi}\end{tabular} & 1 & MSE & DGL & GLN  & \begin{tabular}[c]{@{}l@{}} Intrinsic state probabilities\\Intrinsitic transition probability\\Inverting logic\\Pin state, output state corr. (edge)\\Pin transition to output-pin\\transition corr. (edge)\end{tabular} & NA & NA \\\hline

 \textbf{\cite{gnntesting}}    & Node classification & GCN~\cite{kipf2016semi} & 3&Cross-entropy & Directed & GLN & \begin{tabular}[c]{@{}l@{}}Logic level\\Controllability-0/1\\Observability\end{tabular} & NA & NA \\\hline

\textbf{PL-GNN~\cite{lu2020vlsi,plgnn}} & Node clustering & GraphSAGE~\cite{hamilton2017inductive} & 1 & Unsupervised & Undirected & GLN &Hierarchy and memory features & NA & NA \\\hline

\textbf{\cite{agnesina2020general,9256814}} & GNN-RL & GraphSAGE~\cite{hamilton2017inductive} & - & Unsupervised & Directed & GLN &
\begin{tabular}[c]{@{}l@{}} \#Strongly conn. comp., Gate type\\ Fanout, Area, Max. clique, k-colorability\end{tabular} & - & \begin{tabular}[c]{@{}l@{}}Shallow\\ aggregation\end{tabular}\\\hline

\textbf{\cite{guo22}} & \begin{tabular}[c]{@{}l@{}} Node classification\\
Node regression\\
Edge regression\end{tabular} & Custom~\cite{guo22} & 3 & Custom $L_2$ &Heterogeneous&GLN & \begin{tabular}[c]{@{}l@{}} Fanin/Fanout, PI/PO\\
Pin capacitance, Distance to die boundaries\\
Net distance, LUT indices, valid (edge)\end{tabular} & NA & NA \\\hline

\textbf{DoomedRun~\cite{DoomedRun}}    & Graph regression & GraphSAGE~\cite{hamilton2017inductive} & 2& MSE & - & GLN & \begin{tabular}[c]{@{}l@{}}Worst slack/output slew\\Worst input slew\\Switching, internal, leakage power \end{tabular} & Mean pooling & LSTM\\\hline

 \textbf{DeepPR~\cite{DeepPR}}    & GNN-RL & GCN~\cite{kipf2016semi} & 2 & MSE & Hypergraph & GLN & - & NA & NA \\ \hline

\textbf{ParaGraph~\cite{ren2020paragraph,extparagraph}}   & Node regression & \begin{tabular}[c]{@{}l@{}}GraphSAGE~\cite{hamilton2017inductive}\\GAT~\cite{velivckovic2017graph}\end{tabular} & 2 & MSE & Heterogeneous  & \begin{tabular}[c]{@{}l@{}}Pre-layout\\transistor\\netlist (TNL)\end{tabular}& \begin{tabular}[c]{@{}l@{}}Gate poly length\\ \#Fingers, fins, copies\\ Length of resistor, \#Copies (capac.)\\ Transistor type, Fanout (net)\end{tabular} & NA &NA \\ \hline

\textbf{Circuit-CNN~\cite{zhang2019circuit}}& Graph regression & Custom~\cite{zhang2019circuit} & - & $L_1$ & - &
\begin{tabular}[c]{@{}l@{}}Resonant\\circuit\end{tabular} & \begin{tabular}[c]{@{}l@{}}Side length, Angular slit
position\\Relative position (edge)\\Angular positions (edge)\\Gap and shift length (edge)\end{tabular} &
\begin{tabular}[c]{@{}l@{}}Selects two \\special nodes\\conn. to ports\end{tabular} & Concat \\\hline

\textbf{Circuit Designer~\cite{circuitdesigner}} & GNN-RL & GCN~\cite{kipf2016semi} & 7 & Custom~\cite{circuitdesigner} & Undirected & Transistor &\begin{tabular}[c]{@{}l@{}} Transistor model parameters\\Node type (NMOS, PMOS, R, C)\end{tabular} &NA&NA \\ \hline

\textbf{IronMan~\cite{ironman,wu2021ironman}} & GNN-RL &GCN~\cite{kipf2016semi}&1-4 & \begin{tabular}[c]{@{}l@{}}Mean squared logarithmic\\
MAE\end{tabular}
& DFG &\begin{tabular}[c]{@{}l@{}}HLS C/C++\\ Program\end{tabular} & \begin{tabular}[c]{@{}l@{}}Node types, Data precision \\HLS directive \#pramga\end{tabular}&NA&Mean \\\hline

\end{tabular}%
}
\end{table*}

\subsubsection{Behavioral and Logic Design}

Wu \textit{et~al.}~\cite{ironman,wu2021ironman} tackle design-space exploration (DSE) for HLS, considering user constraints and/or Pareto
optimization.
The framework consists of three modules:
a GNN-based predictor for performance and resource utilization, an RL-based multi-objective DSE engine, and a code
transformer to assist the other modules, by extracting data-flow graphs (DFGs) from high-level C/C++ code and
generating RTL code.

D-SAGE~\cite{ustun2020accurate} covers operation mapping in HLS for FPGAs.
The authors find that structures around arithmetic operations impact the mapping between operations and FPGA
resources and thus timing. Their framework then learns such mapping behaviour to perform mapping-aware timing
characterization of arithmetic modules.

Lopera \textit{et~al.}~\cite{lopera22_ICCAD}
utilize GNNs for timing estimation at RTL.
Their work applies GNNs for predicting delay and slew values of
RTL components, by mapping design components to DAGs with multidimensional node and edge features.
Two GNN architectures are used, MPNN and GCN, and
their extracted node embeddings are passed through an MLP, which performs an edge-level regression task.
With such edge-level prediction, however, this work does not predict critical paths of a design
directly using GNN-based models of circuits, which we expect to be more powerful (Sec.~\ref{sec:GNN_HW_RE}).

\subsubsection{Logic Synthesis}

Kirby \textit{et~al.}~\cite{kirby2019congestionnet} propose the use of GATs for prediction of routing congestion before placement.
Graphs are simple undirected circuit representations with nodes representing gates and edges wires.
Initial features are cell type, function, geometry. Their work is one of the first to use GNNs for EDA tasks.

Ghose \textit{et~al.}~\cite{ghose21} predict routing congestion during synthesis.
Their method works on netlist structures/graphs and can generalize across graphs.
The authors find that popular embedding methods did not perform well for the task, but 
matrix factorization with clustering showed good promise. They also found that deep, wide GATs are not
performing as well as shallow, wide versions of GraphSAGE.

Zhou \textit{et~al.}~\cite{zhou22} propose a heterogeneous, directed GNN model to encode timing graphs.
The model is trained using imitation learning, and aims to accelerate Lagrangian relaxation (LR)-based gate sizing.
The model is shown to transfer well to larger unseen graphs.

Wu \textit{et~al.}~\cite{LOSTIN} propose a hybrid framework to predict quality-of-result. The framework
jointly considers circuit structures
and temporal information, i.e., the order of synthesis steps.
Thus, two hybrid GNN-based models that exploit spatio-temporal information are proposed;
the first uses a GNN to characterize designs via ``supernodes'' encoding temporal
information, the second uses a GNN for
spatial learning and an LSTM for temporal learning.

\subsubsection{Partitioning, Floorplanning}

Lu \textit{et~al.}~\cite{tpgnn} tackle partitioning in 3D ICs, i.e., the assignment of modules to layers in the 3D
stack. First, their work proposes a hierarchy-aware edge-contraction algorithm (to manage 
routing overheads observed with classical bin-based partitioning). Then, layer partitioning/assignment is
mapped to a problem of weighted k-means clustering, which is solved using an unsupervised GNN framework.
Aside from the hierarchy, the feature vectors also describe timing parameters.

Mirhoseini \textit{et~al.}~\cite{mirhoseini2021graph} tackle floorplanning in an industrial context.
The authors approach this problem via RL and develop an edge-based GCN that describes the
high-level netlist of modules.

\subsubsection{Placement and Routing}

Xie \textit{et~al.}~\cite{net2} propose a GAT framework for estimating net lengths prior to placement.
Features are derived from partitioning, e.g., cuts and cluster assignment, and are used as directional edge features to
describe relations between nodes/nets. Hypergraphs
with cell nodes are considered as well.

Lu \textit{et~al.}~\cite{lu2020vlsi,plgnn} present a GraphSAGE-based framework that generates cell clusters to guide placement.
The weighted k-means clustering is derived from learned node embeddings, which build on hierarchy information and
memory affinity as initial features.

Agnesina \textit{et~al.}~\cite{9256814} propose an RL framework for tuning placement, with GraphSAGE used for processing the
netlist graph considering initial features like area, fanout, gate type, etc. More features, describing the netlist
from global perspective, are considered for the overall RL framework. The authors present a similar approach
in~\cite{agnesina2020general} where they also touch upon partitioning in 3D ICs.

Guo \textit{et~al.}~\cite{guo22} tackle the prediction of arrival times and slacks, to support timing-driven placement without
the need for static timing analysis (STA).
Circuits are represented as heterogeneous graphs with edges for nets as well as cells and nodes for cell pins.
Considered features are inspired by timing engines, e.g., pin capacitances and net distances are considered.
The authors argue that ``brute-force stacks'' of (overly) deep GCN layers do not perform well; careful exploration of
the GNN architecture is called for.

Baek \textit{et~al.}~\cite{baek22} predict hotspots for design rule check (DRC) issues during placement, considering both pin access and
routing congestion.
To do so, the authors first devise a pin proximity graph, which models the spatial information on cell pins and
pin-to-pin disturbances/conflicts. Then, they propose an ML model that combines GNN with U-Net; the latter is a fully
convolutional network that is effective for semantic segmentation.
The GNN and U-Net accommodate different features: the GNN uses the pin proximity graph, which captures local
pin accessibility, whereas the U-net considers a set of grid-based features to describe both local and global
perspectives for routing congestion.

Cheng \textit{et~al.}~\cite{DeepPR} present a joint RL framework with gradient-based optimization to tackle both placement and
routing. The authors utilize a multi-view embedding model to encode both global graph level and local node level
information as well as distillation, all to improve exploration of the search space.

\subsubsection{Timing Closure}

Lu \textit{et~al.}~\cite{lu21} present an RL framework with GNN for gate sizing.
Among others, the authors find that the GNN subgraph extraction is essential for predicting the total negative slack (TNS) impact of
the related sub-circuits on the overall circuit.

Lu \textit{et~al.}~\cite{DoomedRun} seek to identify design runs that would fail timing closure, by predicting the
post-route TNS.
Toward this end, the authors leverage GNNs to represent netlist graphs extract from
prior stages, namely placement and clock delivery.
They also utilize LSTM networks to realize sequential modeling of the
design flow.

\subsubsection{Verification, Testing}

Zhang \textit{et~al.}~\cite{GRANNITE} propose supervised learning of toggle rates
for power estimation without the need for gate-level simulations.
The authors represent netlists as graphs, and use register states and inputs from RTL simulation as
features as well as combinational gates' toggle rates as labels.

Ma \textit{et~al.}~\cite{gnntesting} propose a multi-stage GCN model to predict candidates for observation points in a netlist.
Given a set of netlists
with all nodes labeled as either difficult-to-observe or easy-to-observe, a classifier is trained
to find a set of locations where the observation points should be inserted, which
can maximize fault coverage and minimize observation points number and test pattern number.

\subsubsection{Analog, Mixed Signal, and Transistor Design}

Zhang \textit{et~al.}~\cite{zhang2019circuit} use GNNs to tackle distributed circuit design, i.e., analog and mixed-signal
(AMS) circuits operating at high frequencies
with wavelengths comparable to or smaller than circuit components, as seen in 5G and 6G electronics.
The authors show that their GNN model applies for both simulating such circuits as well as automating the
otherwise complex and expert-driven design process. Simulation times are reduced significantly, and new
structures/templates have been found during GNN-based design.

Ren \textit{et~al.}~\cite{ren2020paragraph} predict layout parasitics and device parameters of AMS
circuits, to render pre-layout simulations more efficient and accurate.  The authors convert circuit schematics to
graphs and leverage GraphSage, Relation GCN, and GAT networks.  They argue that such ensemble modeling increases model
accuracy over a large range of prediction values.
Liu \textit{et~al.}~\cite{extparagraph} extend the work, by considering
dropout as an efficient prediction of uncertainty for Bayesian
optimization to automate transistor sizing.
The authors argue that the inclusion of parasitic prediction in
the optimization loop could guarantee satisfaction of all design
constraints, while schematic-only optimization fail numerous
constraints if verified with parasitic estimations.

Wang \textit{et~al.}~\cite{circuitdesigner} propose an RL framework with GCN for automatic transistor sizing. The authors show
that their model enables transfer learning across five nodes and two circuit topologies with superior figure of merit
over prior art.

Chen \textit{et~al.}~\cite{9586211} tackle symmetry matching in AMS circuits.
The authors propose an unsupervised, inductive GNN framework that supports both system-level and device-level symmetry
constraints extraction and is generalizable.  A heterogeneous multi-graph representation has been proposed for interconnection modeling,
and a circuit feature embedding algorithm has been shown
to represent circuits with the most representative substructures.

\subsection{GNNs for Hardware Reliability}
\label{sec:GNN_HW_RE}

Only recently, GNNs have been used to tackle reliability problems. We review selected works below and summarize in
Table~\ref{tab:GNN_RE}.

\begin{table*}[tb]
\caption{Summary of selected GNN frameworks for hardware reliability. NA means not applicable.}
\smallerspacecaption
\label{tab:GNN_RE}
\resizebox{0.95\textwidth}{!}{%
\small
\begin{tabular}{llllllllll}
\hline
\textbf{Platform} & \textbf{GNN Task} & \textbf{GNN} & \textbf{\#Layers} & \textbf{Loss Function} & \textbf{Graph Type} & \textbf{Circuit Type} & \textbf{Features} & \textbf{Pooling} & \textbf{Readout} \\ \hline
\textbf{\cite{AnalogRE}} & Node regression & H-GCN~\cite{AnalogRE} & 2 & MSE & Heterogeneous directed & Post-layout TNL & Design parameters & NA & NA\\ \hline
\textbf{Deep H-GCN~\cite{HGCN}}& Node regression & H-GCN~\cite{HGCN} & 2-8 & MSE & Heterogeneous directed& post-layout TNL  & Design parameters  & NA & NA\\ \hline
\textbf{GNN4REL~\cite{gnn4rel}} & Subgraph regression & PNA~\cite{corso2020principal} & 4 & MSE & Directed & GLN & Gate type &NA & Add\\\hline

\end{tabular}%
}
\end{table*}

Alrahis \textit{et~al.}~\cite{gnn4rel} present a GNN model to predict delay degradations in digital ICs,
for any given timing path, due to process variations and device aging. The objective is
to allow a designer to size guardbands such that correct design operation is ensured, without need for STA nor
for access to sensitive transistor aging models during inference.
The framework extracts subgraphs around timing paths of interest and performs a subgraph regression problem.

Chen \textit{et~al.}~\cite{AnalogRE,HGCN} propose a GCN framework to predict aging-induced transistor degradation in analog
ICs.  Since analog circuits contain various different components (i.e., transistors, capacitors, resistors, etc.), the
authors utilize a heterogeneous graph representation and GNN architecture to tackle this task.

\smallerspacecaption
\subsection{GNNs for Hardware Security}
\label{sec:GNN_HW_SEC}

GNNs have been used to address various hardware security concerns. We review selected works below and summarize in
Table~\ref{tab:my-table}.

\begin{table*}[tb]
\caption{Summary of selected GNN frameworks for hardware security. $-$ means unspecified, and NA means not applicable.}
\smallerspacecaption
\label{tab:my-table}
\resizebox{\textwidth}{!}{%
\small
\begin{tabular}{llllllllll}
\hline
\textbf{Platform} & \textbf{GNN Task} & \textbf{GNN} & \textbf{\#Layers} & \textbf{Loss Function} & \textbf{Graph Type} & \textbf{Circuit Type} & \textbf{Features} & \textbf{Pooling} & \textbf{Readout} \\ \hline
\textbf{GNN4TJ~\cite{gnn4tj}} & Graph classification & GCN~\cite{kipf2016semi} & 2 & Cross-entropy & DFG & RTL/GLN & Type of node oper. & Top-k filtering. & Max \\ \hline
\textbf{GNN4IP~\cite{gnn4ip}} & Graph similarity & GCN~\cite{kipf2016semi} & 2 & Cosine-similarity & DFG & RTL/GLN & Type of node oper. & Top-k filtering & Max \\ \hline
\begin{tabular}[c]{@{}l@{}}
\textbf{GNN-RE~\cite{gnnre}}\\
\textbf{AppGNN~\cite{bucher2022appgnn}}
\end{tabular} & Node classification & GraphSAINT~\cite{zeng2019graphsaint} & 5 & Cross-entropy & Undirected & GLN & \begin{tabular}[c]{@{}l@{}}Input/Output degree\\ Connectivity to ports\\ Gate type\\Neighborhood info\end{tabular} & Multi-head attention & NA \\ \hline

\textbf{ABGNN~\cite{he2021graph}} & Node classification & ABGNN~\cite{he2021graph}& 1-5-2 & Cross-entropy & Directed & GLN & $-$ & NA & NA \\\hline

\textbf{\cite{zhao2022graph}} & Node classification & GCN~\cite{kipf2016semi} & 3 & $-$ & DAG & GLN & \begin{tabular}[c]{@{}l@{}}GNN-RE~\cite{gnnre} features\\Truth table\\Signal probability\\NPN class\end{tabular}& NA & NA \\\hline

\textbf{NCL~\cite{wang2022functionality}} & \begin{tabular}[c]{@{}l@{}}Node classification\\Graph classification\end{tabular} & FGNN~\cite{wang2022functionality} & $-$ & Cross-entropy & Directed & GLN & $-$ & NA & Mean \\\hline

\textbf{ReIGNN~\cite{chowdhury2021reignn}} & Node classification & GraphSAGE~\cite{hamilton2017inductive} & 3 & 
\begin{tabular}[c]{@{}l@{}}Negative \\ log-likelihood\end{tabular}

 & $-$ & GLN & \begin{tabular}[c]{@{}l@{}}Input/Output degree\\ Harmonic centrality\\ Gate type \\ Betweenness centrality\\ Neighborhood info\end{tabular} & NA & NA \\\hline

\textbf{ICNet~\cite{chen2020estimating}} & Graph regression & GAT~\cite{velivckovic2017graph} & 2 & \begin{tabular}[c]{@{}l@{}}Mean squared\\ error (MSE)\end{tabular} & Undirected&GLN & \begin{tabular}[c]{@{}l@{}} Gate type\\
Key-gate mask\end{tabular} & Attention & Attention \\\hline

\begin{tabular}[c]{@{}l@{}}
\textbf{GNNUnlock~\cite{alrahis2020gnnunlock}}\\
 \textbf{GNNUnlock+~\cite{gnnunlockp}}
\end{tabular}
 & Node classification & GraphSAINT~\cite{zeng2019graphsaint} & 2 & Cross-entropy & Undirected & GLN & \begin{tabular}[c]{@{}l@{}}Input/Output degree\\ Connectivity to ports\\ Gate type\end{tabular} & NA & NA \\ \hline
\textbf{OMLA~\cite{omla}} & Subgraph classification & GIN~\cite{xu2018powerful} & 5-6 & Cross-entropy & Undirected & GLN & \begin{tabular}[c]{@{}l@{}}Gate type\\ Connectivity to ports\\ Distance encoding\end{tabular} & \begin{tabular}[c]{@{}l@{}}Summing node features\\ from the same layer\end{tabular} & Sum \\ \hline

\begin{tabular}[c]{@{}l@{}}
\textbf{MuxLink~\cite{muxlink}}\\
\textbf{UNTANGLE~\cite{untangle}} 
\end{tabular}
& Link prediction & DGCNN~\cite{zhang2018end} & 4 & Cross-entropy & Undirected & GLN & \begin{tabular}[c]{@{}l@{}}Gate type\\ Distance encoding\end{tabular} & DGCNN sortpooling~\cite{zhang2018end} & Max \\ \hline

\textbf{Titan~\cite{titan}} & Subgraph classification & GIN~\cite{xu2018powerful} & 6 & Cross-entropy & Undirected & GLN &\begin{tabular}[c]{@{}l@{}}Gate type, \#Inputs\\ Distance encoding\end{tabular} & \begin{tabular}[c]{@{}l@{}} Summing node features\\from the same layer \end{tabular} & Sum\\\hline
\end{tabular}%
}
\end{table*}

\subsubsection{HT Detection}
\label{sec:GNNs_HT_detection}

Yasaei \textit{et~al.}~\cite{gnn4tj} propose GNN4TJ,
a GNN platform for HT detection
without prior knowledge of the design IP or the HT structure. 
GNN4TJ first converts a set of RTL designs into their corresponding DFGs, which are
fed to a GNN to learn about the structure
of the designs.
Then, given a design to test, the GNN performs graph classification to predict the presence/absence of HTs across the entire design.
Other works extend the concept of this framework using 
GLNs~\cite{GNN4TJ_Journal,yu2021hw2vec}.

Further GNN-based platforms~\cite{muralidhar2021contrastive,yasaei2022golden} importantly
enable HT localization, i.e., HT detection at the node-/gate-level.

\subsubsection{Piracy Detection}
\label{sec:background_IP}

Yasaei \textit{et~al.}~\cite{gnn4ip} propose a GNN-based method for detecting IP piracy. The framework, GNN4IP, assesses the similarity
between the original and the questionable circuit to predict theft.
GNN4IP can assess circuits given either as RTL or GLN.
GNN4IP does not require the addition of any watermarks or fingerprints, reducing overheads and avoiding removal attacks~\cite{alkabani2007remote,cui2015ultra}.
This is because the GNN generates an embedding for each circuit from its underlying structure, which is considered as ``signature'' of
the design.
Subsequently, the GNN optimizes the embeddings such that distances in the embedding space reflect the similarity
between designs~\cite{hamilton2017inductive}.
Finally, GNN4IP computes the cosine similarity score for embeddings 
to predict piracy across the two circuits.

Yu \textit{et~al.}~\cite{yu2021hw2vec} propose HW2VEC, which combines GNN4TJ~\cite{gnn4tj} and GNN4IP~\cite{gnn4ip} into a
single framework.
   HW2VEC supports different graph types, such as DFG and abstract syntax tree (AST).

\subsubsection{Functional Reverse Engineering}

Alrahis \textit{et~al.}~\cite{gnnre} propose GNN-RE, a GNN-based platform for functional RE.
A given GLN is first transformed into an undirected graph, with
gates represented as nodes that are initialized with a feature vector of gate type,
neighboring gates, input degree, output degree, and connectivity to primary ports.
Then, GNN-RE performs node classification to identify which gates belong to which sub-circuit/module.

In ReIGNN~\cite{chowdhury2021reignn}, GLNs are represented as graphs which the GNN processes to
discriminate between state and data registers.

He \textit{et~al.}~\cite{he2021graph} propose an asynchronous bidirectional GNN (ABGNN) for sub-circuit classification in GLNs, focusing on
arithmetic blocks and specifically adders.
ABGNN represents GLNs as directed graphs, to maintain the natural data flow direction of circuits.
ABGNN predicts boundaries of adder blocks by training two
separate GNNs;
one aggregates information from
the respective fan-in cones and the other from the fan-out cones. The two generated embeddings are then combined as final embedding.

The above methods focus solely on structural circuit properties for prediction.
In contrast, Wang \textit{et~al.}~\cite{wang2022functionality} propose a contrastive learning (CL)-based framework for netlist
representation, to extract the logic functionality, which is universal and transferable across different circuits.
The authors also propose a functionality-centric GNN that works in tandem with the netlist representation framework classification purposes.
This GNN builds on ABGNN~\cite{he2021graph}, but uses independent aggregators for different node types.

Zhao \textit{et~al.}~\cite{zhao2022graph} employ a GNN for functional RE under circuit
rewriting.  Authors consider the features used for GNN-RE~\cite{gnnre} and a set of functionality-aware
features, thus improving the prediction accuracy. Features are based on truth table information,
signal probability, and negation-permutation-negation (NPN) class.

Bucher \textit{et~al.}~\cite{bucher2022appgnn} propose AppGNN, a platform that extends on GNN-RE~\cite{gnnre} to
handle approximate circuits. AppGNN employs a node-sampling algorithm while training the GNN on exact circuits,
allowing the model to generalize to approximate computing.

Kunal \textit{et~al.}~\cite{gana} propose GANA, a GCN-based platform for RE of analog circuits.

\subsubsection{Attacks on Design-for-Trust Methods}

Various GNN-based algorithms have demonstrated strong results for breaking logic locking and layout camouflaging,
either by learning on and predicting the structures of the IC netlists despite the presence of these
design-for-trust measures, or by directly learning, predicting, and bypassing the structures of these
measures themselves~\cite{ml_locking_survey,gnnunlockp,omla,untangle,muxlink}. 
The related GNN attacks work predominantly in an oracle-less
setting, which is more powerful than seminal attacks like the Boolean Satisfiability (SAT)-based attack~\cite{Subramanyan_host_2015}. 
These SOTA GNN works exposed the following shortcomings in the design-for-trust measures.

\textbf{Key leakage.}
OMLA is a GNN-based attack on X(N)OR locking~\cite{omla}. OMLA represents the locked GLN as an undirected graph
and extracts an h-hop enclosing subgraph around each key-gate. The subgraphs include information on the key-gates
themselves but also on their surrounding circuitry. Thus, OMLA deciphers the key-bit values by performing subgraph classification.

\textbf{Link formation.} 
MuxLink~\cite{muxlink} and UNTANGLE~\cite{untangle} exposed a new vulnerability -- link formation -- to break
MUX-based logic locking~\cite{sisejkovic2021deceptive,alaql2021scope,InterLock} which was previously considered
ML-resilient.
The intuition behind MuxLink and UNTANGLE is that modern ICs contain a considerable degree of repetition and IP reuse~\cite{saha2011soc}. 
Thus, MuxLink and UNTANGLE both employ a GNN to learn the remaining/unlocked parts of a locked design and then perform
predictions regarding the obfuscated interconnects. 
More specifically, MuxLink and UNTANGLE convert the problem of deciphering the inputs of a MUX key-gate to a
link-prediction problem and solves it using a GNN. Titan~\cite{titan} is a GNN-based attack on large-scale obfuscation
that combines the concepts of OMLA, MuxLink, and UNTANGLE into a holistic attack platform that can handle both gate and
interconnect obfuscation. To achieve this, Titan expands the GNN model to more
classes and considers more target nodes.

\textbf{Structural leakage.} GNNUnlock~\cite{gnnunlockp,alrahis2020gnnunlock} targets on provably-secure logic
locking (PSLL). It use a GNN to identify and isolate the protection logic in PSLL, which is
generally embedded in the design. Identifying this protection logic facilitates its removal, enabling the
recovery of the original design~\cite{removal,limaye2022valkyrie}.

\textbf{Attack scalability.} As indicated, the SAT-based attack is a seminal attack on logic locking~\cite{Subramanyan_host_2015}. 
Although this attack succeeds on a wide range of locking techniques, its run-time can vary from seconds to days,
	 depending on the circuit and the locking technique. 
The ICNet~\cite{chen2020estimating} platform represents the locked netlist as graph and encodes each
node/gate with a key-gate mask, indicating if the node represents a key-gate or not, and the gate type.
The employed GCN then performs a regression task, predicting the run-time of the SAT-based attack on the given circuit.

\section{Conclusions}
\label{sec:conclusions}

We have provided a comprehensive survey of GNNs used for different domains of IC design. We classify
selected SOTA works into three categories; GNNs for EDA, GNNs for reliable ICs, and GNNs for hardware security.
The references, data sets, codes, and resources are collected and summarized in our GNN4IC hub. 

Although GNNs are natural candidates for learning on circuits and have shown excellent promises,
employing them comes with unique challenges.
For example, some works had to resort to custom GNN architectures and edge/node attributes/weights to achieve
reasonable expressiveness of the model while still maintaining the directionality of circuits~\cite{ustun2020accurate,gnntesting,gnnre}.
Other works have devised heterogeneous GNN architectures to be able to handle analog circuits with their different
types of elements/nodes.
Prior works largely focus on the structure of circuits, losing out on functional properties. Most
recent works, however, have shown that incorporating functional attributes can enhance the
performance of the GNN platforms~\cite{zhao2022graph,wang2022functionality}. Researchers have also focused on scalability for
training of different GNN models: most prior works implement some form of graph sampling or
clustering to speed up the training process~\cite{gnnre,gnnunlockp}.
More specific challenges and solutions are, among others, devising custom matrix multiplication 
to avoid repetitive computations~\cite{gnntesting} or mimicking design rules into the model~\cite{zhang2019circuit}.

Across the different domains covered in this survey, GNNs for hardware reliability represents a relatively new
direction, which we expect to grow in near future.  The direction of using GNNs for EDA is more mature and
well-investigated, considering different graph types, architectures, node features, and downstream tasks.  As for
hardware security, we note that researchers have used GNNs mainly to evaluate security promises of various
design-for-trust methods. But, GNNs are yet to be explored for their potential to render IC design secure.  In
general, when using GNNs for these various critical tasks, we expect the concepts of trustworthy ML and explainable AI
to have a positive impact on the field in future.



\end{document}